\documentclass[runningheads]{llncs}
\usepackage{amsmath}\usepackage{amssymb}

\usepackage[T1]{fontenc}
\usepackage{graphicx}
\usepackage{tikz}
\usepackage{comment}
\usepackage{orcidlink}
\usepackage{booktabs}
\usepackage{multirow}
\usepackage{adjustbox}
\usepackage{microtype}

\let\svthefootnote\thefootnote
\newcommand\freefootnote[1]{%
  \let\thefootnote\relax%
  \footnotetext{#1}%
  \let\thefootnote\svthefootnote%
}
 
\global\long\def\modelshort{\textrm{{CATS}}}
\usepackage{color}

\begin{document}
\title{From Category to Scenery: An End-to-End Framework for Multi-Person Human-Object Interaction Recognition in Videos}

\titlerunning{From Category to Scenery}
%
\author{Tanqiu Qiao\orcidlink{0000-0002-6548-0514} \and
Ruochen Li\orcidlink{0000-0001-8966-9613} \and
Frederick W. B. Li\orcidlink{0000-0002-4283-4228}\index{Li, Frederick W. B.} \and
Hubert P. H. Shum\orcidlink{0000-0001-5651-6039}}\index{Shum, Hubert P. H.} 
\authorrunning{Qiao et al.}
\institute{Durham University, Durham, United Kingdom\\
\email{\{tanqiu.qiao, ruochen.li, frederick.li, hubert.shum\}@durham.ac.uk}}
\maketitle              
\begin{abstract}
Video-based Human-Object Interaction (HOI) recognition explores the intricate dynamics between humans and objects, which are essential for a comprehensive understanding of human behavior and intentions. While previous work has made significant strides, effectively integrating geometric and visual features to model dynamic relationships between humans and objects in a graph framework remains a challenge. In this work, we propose a novel end-to-end category to scenery framework, $\modelshort$, starting by generating geometric features for various categories through graphs respectively, then fusing them with corresponding visual features. Subsequently, we construct a scenery interactive graph with these enhanced geometric-visual features as nodes to learn the relationships among human and object categories. This methodological advance facilitates a deeper, more structured comprehension of interactions, bridging category-specific insights with broad scenery dynamics. Our method demonstrates state-of-the-art performance on two pivotal HOI benchmarks, including the MPHOI-72 dataset for multi-person HOIs and the single-person HOI CAD-120 dataset.

\end{abstract}
\section{Introduction}
Human-Object Interaction (HOI) recognition delves into the subtle dynamics between humans and objects, aiming to capture the breadth of their interactions from basic actions to complex activities. This field transcends mere identification to explore the depth of their interactions, from elementary actions to intricate sequences, which are essential for a comprehensive understanding of human behavior and intentions \cite{morais2021learning,qiao2022geometric,zhuo2019explainable}. Accurate HOI recognition is crucial across various domains, serving as a cornerstone for developing sophisticated surveillance \cite{dogariu2020human,rezaee2024survey}, enhancing video analysis techniques \cite{nagarajan2019grounded,liu2020forecasting,li2022multiclass_sgcn}, and facilitating effective human-robot collaboration \cite{smith2013gaze,mukherjee2022survey}.

Prior work in Human-Object Interaction (HOI) detection predominantly examines interactions within static images, offering crucial insights yet constrained by the lack of temporal dynamics \cite{gkioxari2015actions,mallya2016learning,gao2018ican}. The emergence of single-person HOI video datasets marks a significant advancement \cite{koppula2013learning,dreher2020learning,KrebsMeixner2021}, enabling the development of models that understand spatio-temporal actions through visual cues \cite{qi2018learning,jain2016structural,morais2021learning}. A notable progression is presented by \cite{qiao2022geometric}, which leverages geometric features informed networks for HOI recognition in videos, broadening the scope to encompass two-person HOIs with the introduction of a novel dataset.

\textls[-3]{While fusing geometric and visual features achieves remarkable performance, video-based HOI recognition still faces challenges in effectively fusing these features and learning dynamic relationships between humans and objects in a graph model. 2G-GCN \cite{qiao2022geometric} attempts to enrich visual data with geometric information via a graph-based network. However, merging geometric features of all humans and objects with individual visual features in a single graph leads to a critical flaw by neglecting category-specific characteristics. This fusion difficulty hampers accurate and specific HOI learning, especially in complex multi-person scenes.}

Categorization simplifies learning and improves behavior discrimination by grouping similar features, enhancing model accuracy in identifying diverse interactions. In this work, we follow natural cognitive processes \cite{li2020nips,baldassano2017human} to learn HOIs from category-level feature fusion to scenery-level graph representation, facilitating a structured and comprehensive understanding. This strategy enables a more sophisticated integration of varied feature types, ensuring each level is fully leveraged for enhanced representational efficacy. We propose a novel end-to-end \underline{CAT}egory to \underline{S}cenery framework ($\modelshort$), which initially generates geometric features via a graph for different categories, integrating them with corresponding visual features. Subsequently, a scenery interactive graph is constructed using these enriched geometric-visual features as nodes, to deeply understand the interaction dynamics among all humans and objects.

Our approach surpasses state-of-the-art performance on two HOI benchmarks, including the two-person MPHOI-72 \cite{qiao2022geometric} dataset and the single-person HOI CAD-120 \cite{koppula2013learning} dataset. Additionally, we conduct ablation studies to evaluate the core components of our model. Our main contributions are:

\begin{itemize}
  \item We propose an end-to-end framework $\modelshort$ ranging from category-level feature fusion to scenery-level graph for multi-person HOI recognition in videos.

  \item We propose a multi-category multi-modality fusion module that fuses visual features and graph-based geometric features for human and object categories, respectively.

  \item We propose a scenery interactive graph to learn the relationships among human and object categories via an attention-based graph.
  
\end{itemize}

\section{Related Work}
\subsection{HOI Recognition in Videos}
There are two setups for video-based HOI recognition, where the more challenging setup focuses on segmenting and recognizing distinct human sub-activities in videos. Deep neural networks (DNNs) and graphical models have been combined in recent works. A paradigm for integrating the effectiveness of spatio-temporal graphs with Recurrent Neural Networks (RNNs) in sequence learning is presented by Jain et al. \cite{jain2016structural}. Using learnable graph structures for videos, Qi et al. \cite{qi2018learning} expand previous graphical models in DNNs and pass messages through GPNN. For the intention of acquiring spatial relations, Dabral et al. \cite{dabral2021exploration} compare GCNs to Convolutional Networks and Capsule Networks. In attempting to investigate the evolution of spatio-temporal connections and identify objects in a scene~\cite{li23lim3d,wang2021spatio}, STIGPN \cite{wang2021spatio} utilizes visual-based multi-modal features and a multi-stream fusion strategy to enhance the reasoning capability of the model. Morais et al. \cite{morais2021learning} present a visual feature attention model to learn asynchronous and sparse HOI in videos. Xing et al. \cite{xing2022understanding} represent the 2D or 3D spatial relation of human skeletons and object center points from the detection results in video data as a graph. Based on prior visual-only and geometric-only approaches, 2G-GCN \cite{qiao2022geometric} incorporates geometric features to complement visual features into the HOI recognition network through a graph network. Nevertheless, the fusion of geometric and visual features introduces certain design complexities that offer opportunities for further refinement.

Another more relaxed setup in HOI recognition aims to generate {\texttt{<human, predicate, object>}} triplets, neglecting a more detailed analysis of specific actions and interactions. For example, in recent years, SERVO-HOI \cite{agarwal2023skew} presents a robust end-to-end framework adept at recognizing HOIs within in-the-wild videos, especially effective in high label-skew settings. Zeng et al. \cite{zeng2023cognition} introduce the Relation-Pose Transformer (RPT), a novel framework designed to intricately model the spatial and temporal dynamics between relations and poses, adept at encapsulating spatially contextualized information and the temporal evolution of relationships. Furthermore, Zhang et al. \cite{zhang2023human} explore a new task, Human-Object-Object Interaction (HOOI) detection, focusing on localizing the human and identifying their interactions within untrimmed videos as a quadruple {\texttt{<human, interaction, object1, object2>}}. In this work, our study concentrates on the more challenging aspect of video-based HOI recognition, specifically the segmentation and recognition of distinct human sub-activities along the video timeline.

\subsection{Graph-based HOI Analysis}
Graphical models facilitate the sharing of contextual information among nodes. Qi et al. \cite{qi2018learning} introduce this concept in HOI detection, where they propose a fully-connected graph with detected instances as nodes and update node features with a message passing algorithm. Wang et al. \cite{wang2020contextual} suggest that adaptation to two sets of heterogeneous nodes, human and object, is essential for graph-based HOI analysis. This necessitates modelling intra-class messages differently from inter-class messages during message passing. Incorporating the heterogeneity of nodes, Gao et al. \cite{gao2020drg} create separate human-centric and object-centric graphs for HOI detection by treating human-object pairs as nodes and employing the pairwise spatial relations as node encoding. VSGNet et al. \cite{ulutan2020vsgnet} leverages graph convolution and spatial configuration to refine visual features of human-object pairs and exploits structural connections between them. SCG \cite{zhang2021spatially} develops a bipartite graph to model interrelationships between nodes in HOI scene where each human node is connected to each object node. Building upon SCG, Park et al. \cite{park2023viplo} design a graph with a pose-conditioned self-loop structure to update the encoding of human nodes with local features of skeleton joints. Additionally, Zhang et al. \cite{zhang2022exploring} construct an interaction-centric graph by treating selected interaction proposals as graph nodes to examine inter-interaction semantic structure and intra-interaction spatial structure.

Recent HOI recognition tasks are also inspired by graphical models. LIGHTEN \cite{sunkesula2020lighten} employs a graph structure to model human and object embeddings, which serves them as nodes in the scene. In a similar vein, Dabral et al. \cite{dabral2021exploration} investigate the efficacy of GCNs in spatial relation learning compared to Convolutional Networks and Capsule Networks. Wang et al. \cite{wang2021spatio} propose the STIGPN to understand the evolution of spatio-temporal relationships and distinguish the objects involved in the background using parsed graphs. Xing et al. \cite{xing2022understanding} introduce a novel spatial attention mechanism that can enhance action recognition by adaptively generating a spatial-relation graph during HOIs. In 2G-GCN \cite{qiao2022geometric}, linking collective geometric features with individual visual features causes hierarchical misalignment, as high-level spatial information may not align well with detailed, entity-specific visual data. This focuses on less relevant objects and fails to explicitly learn HOIs. In this study, we develop an understanding of HOIs by progressing from category-level feature fusion to scenery-level graph representation, enabling a structured and thorough comprehension of interactions.

\section{Methodology}
\begin{figure}
\centering
\includegraphics[width=\linewidth]{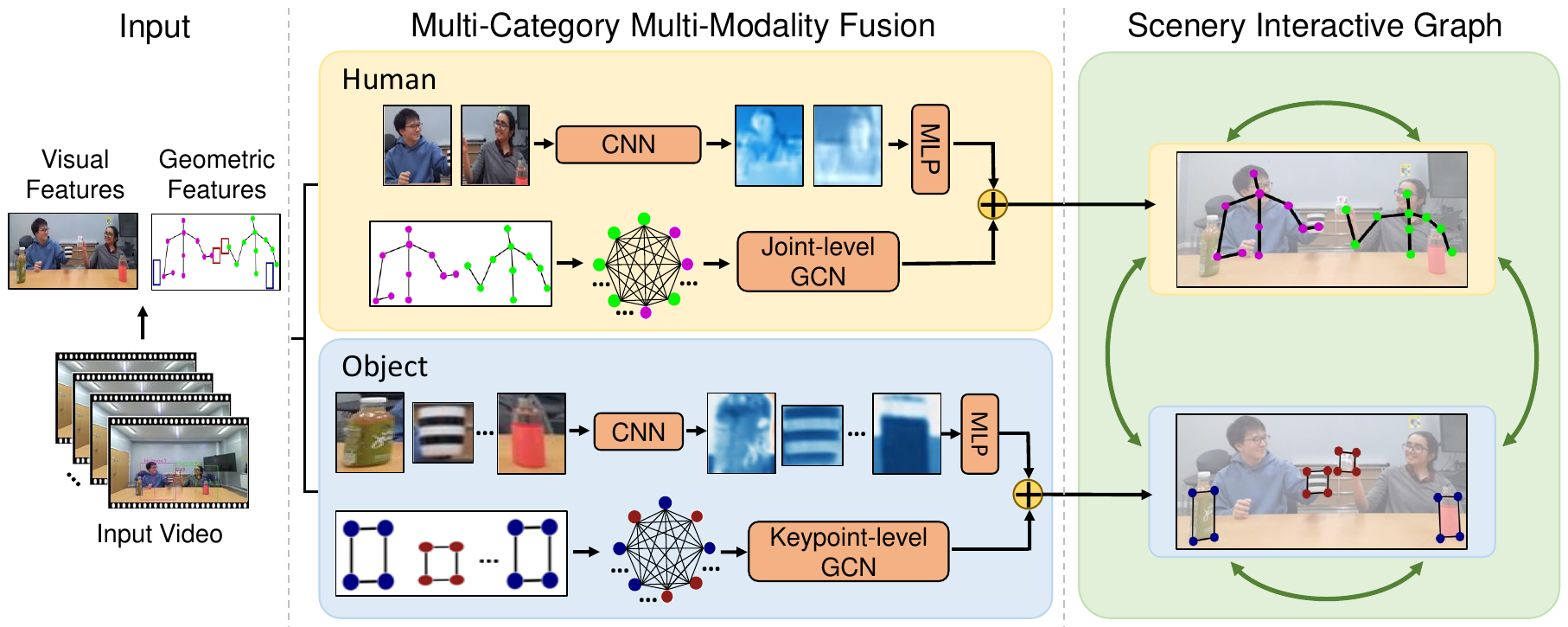}
\caption{Overview of our end-to-end framework $\modelshort$. We first learn geometric features via a graph for human and object categories, fusing them with corresponding visual features. Subsequently, a scenery interactive graph is constructed to deeply understand the interaction dynamics between multi-categories.}
\label{fig:framework}
\end{figure}

We propose an end-to-end framework $\modelshort$ (Fig.~\ref{fig:framework}) to learn HOIs from category-level to scenery-level, which first focuses on the inherent characteristics of different categories, capturing their physical properties and contextual visual cues to achieve a rich feature representation. It then adopts a graph attention neural network to learn multi-category features as a scenery graph representation, which represents the true HOI. This approach mirrors natural cognitive processes \cite{li2020nips,baldassano2017human} facilitating a structured and comprehensive understanding of interactions within various contexts.

Alternative architecture performs suboptimally, an approach treats each human and object as an entity independently, ignoring the correlation between the same category and compromising the model's ability to understand complex dynamics. An alternative method \cite{qiao2022geometric} groups all human poses and object bounding boxes into a single category for geometric feature learning, and then combines these geometric features with visual features in a single graph learning, which complicates entity representation and hampers explicit HOI learning. We compare these alternative architectures with our method in Experimental Results \ref{sec:experiments}.

\subsection{Multi-Category Multi-Modality Fusion}
Previous CNN-based methods for HOI recognition in videos have predominantly focused on visual features \cite{maraghi2019zero,le2020bist,morais2021learning}, which may not be sufficient in cases of occlusion. While more advanced approaches like 2G-GCN \cite{qiao2022geometric} have attempted to incorporate geometric features to complement visual features, they categorize all human skeletons and object bounding boxes under a single category for geometric feature learning, thereby neglecting the distinct characteristics unique to each category and potentially generating skewed geometric features.

To this end, we propose a multi-category multi-modality fusion module that first learns geometric features via a graph for human and object two categories and then fuses them with corresponding visual features (Fig.~\ref{fig:framework}). These category-specific features establish a rich multimodal context, providing a solid foundation for subsequent accurate interaction recognition.

\subsubsection{Geometric Features}
For feature representation in human category and other related tasks, following previous successes \cite{qiao2022geometric,li2024rapidseg}, we concatenate the position and velocity of all humans into keypoint channels, forming human geometric features $\mathcal{HG} = \{hg_{t,h,j}\}_{t=1,h=1,j=1}^{T,H,J} \in \mathbb{R}^4$, where $hg_{t,h,j}$ denotes the body joint of type $j$ in human $h$ at time $t$, $T$ denotes the total number of frames in the video, $H$ and $J$ denote the total number of humans and keypoints of a human body in a frame, respectively. Similar to humans, object geometric features $\mathcal{OG} = \{og_{t,o,u}\}_{t=1,o=1,u=1}^{T,O,2} \in \mathbb{R}^4$, where $og_{t,o,u}$ denotes the bounding box diagonal points $u$ in object $o$ at time $t$ and $O$ denotes the total number of objects.

\begin{figure}
\centering
\includegraphics[width=0.7\linewidth]{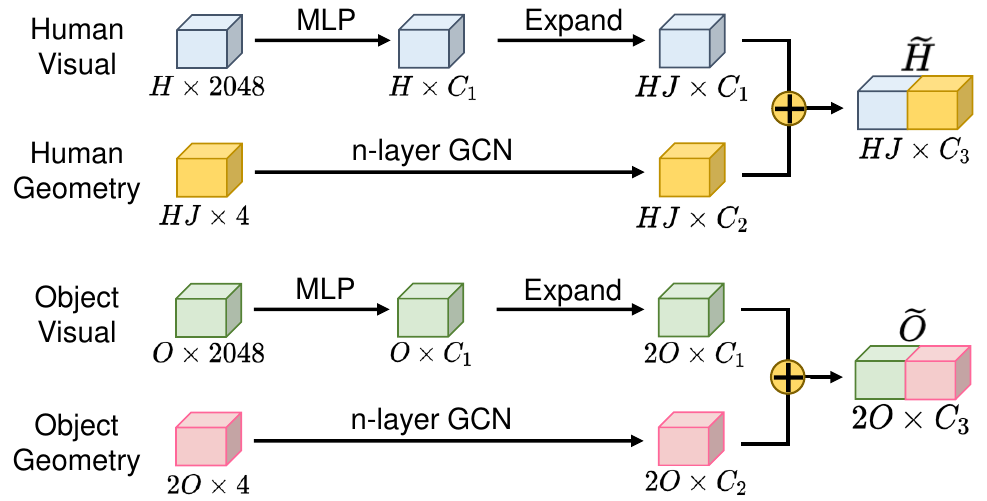}
\caption{The process of learning and fusing geometric and visual features for human and object categories.}
\label{fig:feature_embedding}
\end{figure}

As shown in Fig.~\ref{fig:feature_embedding}, human and object geometric features are adopted n-layer GCNs to capture spatial dynamics and interactions in each category. This enables deeper analysis through successive transformations, allowing the graph-based network to learn intricate patterns of spatial dynamic interactions at multiple levels of abstraction \cite{you2020l2,du2019graph}. Here, taking human geometric features as an example, the operation of each GCN layer is formalized as:

\begin{equation}
H^{(l+1)} = \sigma\left( A H^{(l)} W^{(l)} \right),
\end{equation}
where $H^{(l)}$ represents the activation matrix at the $l$th layer ($H^{(0)} = \mathcal{HG}$ for the initial layer), $A$ is the adjacency matrix defining the graph structure, $W^{(l)}$ is the weight matrix for the $l$th layer, and $\sigma$ is the Tanh activation function.

For an n-layer GCN, this transformation is applied iteratively to obtain the final embedded human geometric features:
\begin{equation}
HG^{\prime} = H^{(n)} = \sigma\left( A H^{(n-1)} W^{(n-1)} \right)
\end{equation}
where $n$ is the total number of GCN layers, iterating the process from $l = 0$ to $n-1$. We choose $n = 4$ based on empirical experimental results. Through this operation, we can obtain the embedded human and object geometric features: $HG^{\prime} \in \mathbb{R}^{T \times HJ \times C_2}$ and $OG^{\prime} \in \mathbb{R}^{T \times 2O \times C_2}$.

\subsubsection{Visual Features}
In contrast to geometric features, visual features in videos offer a wealth of contextual information and essential feature representations. Following \cite{morais2021learning,qiao2022geometric}, we derive 2048-dimensional visual features of entities from Region of Interest (ROI) pooled 2D bounding boxes around humans and objects in video frames. As shown in Fig.~\ref{fig:feature_embedding}, they are subsequently reduced dimensionally to $C_1$ through an MLP with learnable embeddings and aligned dimensionally with geometric features. This process results in the embedded human and object visual features: $HV^{\prime} \in \mathbb{R}^{T \times HJ \times C_1}$ and $OV^{\prime} \in \mathbb{R}^{T \times 2O \times C_1}$.

\subsubsection{Multi-Modality Fusion}

Finally, we fuse embedded geometric and visual features in the human and object keypoint channel, producing new enriched human and object feature representations, respectively: 
\begin{align}
\widetilde{H} &= HG^{\prime} \scalebox{1.2}{$\oplus$} HV^{\prime} \in \mathbb{R}^{T \times HJ \times C_3}; \\
\widetilde{O} &= OG^{\prime} \scalebox{1.2}{$\oplus$} OV^{\prime} \in \mathbb{R}^{T \times 2O \times C_3},
\end{align}
where \scalebox{1.2}{$\oplus$} represents concatenate operation and $C_3 = C_1 + C_2$. This refined fusion of geometric and visual cues creates a richly contextualized blend, laying a solid foundation for enhanced scenery graph learning of HOIs.

\subsection{Scenery Interactive Graph}

To effectively model the interactions between humans and objects, the existing method \cite{morais2021learning} focuses exclusively on their visual features to construct an interaction graph. This approach taps into the visual aspect of interactions, which is essential but insufficient for grasping the dynamic spatial relationships critical to understanding the complexities of HOI. Furthermore, 2G-GCN \cite{qiao2022geometric} offers a more comprehensive view but fuse geometric features representing all entities with visual features representing individuals, which results in hierarchical misalignment and fails to explicitly learn HOIs.

To overcome the constraints of prior approaches, we propose a scenery interactive graph that adopts a graph attention neural network to learn interactions between different categories with enriched feature representation (Fig.~\ref{fig:framework}), to deeply understand the interaction dynamics among all humans and objects. This structured approach facilitates a comprehensive understanding of interactions within various contexts.

\subsubsection{GAT for Learning Scenery Graph}
Specifically, we adopt Graph Attention Networks (GAT) \cite{huang2019stgat} in learning scenery graph interactions is particularly advantageous due to their ability to dynamically adjust to rapid changes in human and object interactions within scenery graphs, thanks to their adaptive edge weighting and handling of non-static features. This ensures a precise focus on relevant entities and their evolving relationships, optimizing the model's responsiveness to the complex dynamics of interactions.

We construct the HOI scenery graph $\mathcal{G}_{s-t} = (\mathcal{V}, \mathcal{E})$, where $\mathcal{V} \in \mathbb{R}^{T \times (HJ + 2O) \times C_{3}}$ represents the node features, which is obtained by concatenating the local human feature representation $\widetilde{H}$ and object feature representation $\widetilde{O}$, and $\mathcal{E} \in \mathbb{R}^{T \times (HJ + 2O) \times (HJ + 2O)}$ denotes the initialized fully-connected adjacency matrix. For each node $\mathcal{V}_i$ at time step $t \in [1, \dots T]$, the feature representation is:
\begin{equation}
\mathcal{V}_{i}^{t} = \sigma\left(\sum_{j \in \mathcal{N}_{(i) \cup { i }}} \alpha_{i,j}^{t} \mathbf{\Theta} \mathcal{V}_{j}^{t}\right),
\end{equation}
and the attention coefficients $\alpha_{i,j}$ are computed as:
\begin{equation}
\alpha_{i,j}^{t} = \frac{\exp\left(\mathrm{LeakyReLU}\left(\mathbf{W}^{\top}[\mathbf{\Theta} \mathcal{V}_{i}^{t}, \Vert , \mathbf{\Theta} \mathcal{V}_{j}^{t}]\right)\right)}{\sum_{n \in \mathcal{N}_{(i) \cup { i }}} \exp\left(\mathrm{LeakyReLU}\left(\mathbf{W}^{\top}[\mathbf{\Theta} \mathcal{V}_{i}^{t}, \Vert , \mathbf{\Theta} \mathcal{V}_{n}^{t}]\right)\right)},
\end{equation}
where $\mathbf{\Theta(\cdot)}$ is the transformation function, $\mathcal{N}(\cdot)$ is the neighbor set of node $i$ and $\mathbf{W}$ represents learnable parameters. This dynamic weighting is crucial as it allows the model to adaptively focus on the most relevant nodes and edges, reflecting the changing nature of interactions and relationships within the scene.

\subsubsection{RNN-based Network for Learning Temporal Dependency}
After obtaining the learned HOI scenery graph representations at each time step $t$, we employ an RNN-based network to learn the temporal dependencies across all the time steps. Specifically, we utilize a Bi-direction Gated Recurrent Unit (Bi-GRU) \cite{chung2014gru} that enables our model to integrate both past and future contexts, enhancing its understanding of the sequential dynamics in human-object interactions. The GRU's gating mechanisms effectively manage long-term dependencies, ensuring robust temporal modeling. For the learned step-wise feature representations, we utilize a Gumbel-Softmax module \cite{jang2016gsm}, enabling precise and adaptable delineation of sub-event lengths in video sequences. This module is instrumental in enabling gradient-based optimization while maintaining probabilistic integrity in segmenting actions, a crucial aspect when dealing with the inherently fluctuating characteristics of video content. Subsequently, we employ another Bi-GRU to discern the temporal relations among segmented sub-actions. The processed features are then leveraged to identify specific sub-activities associated with humans, with the granularity of recognition tailored to suit the requirements of the specific dataset.

\section{Experiments}
\label{sec:experiments}
\subsection{Datasets}
We evaluate $\modelshort$ on two datasets: MPHOI-72 \cite{qiao2022geometric} and CAD-120 \cite{koppula2013learning}, showcasing the superior results on multi-person and single-person HOI recognition.

The MPHOI-72 dataset is valuable for two-person HOI tasks. It contains 72 videos of 8 pairs of people performing 3 distinct activities (\textit{Cheering}, \textit{Hair cutting} and \textit{Co-working}) with 13 human sub-activities (e.g., \textit{Sit}, \textit{Pour}). Each video showcases two participants interacting with 2-4 objects from 3 unique angles. Geometric features and human sub-activities labels are frame-wise annotated.

CAD-120 is a prominent dataset for single-person HOI recognition. It contains 120 RGB-D videos, capturing 10 distinct activities executed by 4 participants, each repeated three times. In each video, a participant interacts with 1-5 objects. The dataset provides frame-wise annotations for 10 human sub-activities (e.g., \textit{opening}, \textit{placing}).

\subsection{Evaluation Protocol}
Following the evaluation protocol of \cite{morais2021learning,qiao2022geometric}, we assess $\modelshort$ across two specific tasks: joint segmentation and label recognition for pre-segmented entities. The initial task involves both segmenting and classifying the timeline of each entity in a video, while the second extends this by assigning labels to pre-segmented sections with known ground truth. We adopt the $\mathrm{F}{1}@k$ metric \cite{lea2017temporal} for evaluation, using standard thresholds of $k=10\%$, $25\%$, and $50\%$. This metric, prevalent in segmentation research \cite{lea2017temporal,farha2019ms,morais2021learning}, determines the correctness of a predicted action segment based on its minimum Intersection over Union (IoU) overlap with the ground truth and is particularly effective for assessing brief actions and detailed segmentation. For dataset evaluation, we implement a leave-two-subjects-out strategy for the MPHOI-72 dataset and a leave-one-subject-out cross-validation approach for CAD-120.

\subsection{Network Setting}
The visual features of humans and objects are extracted from 2D bounding boxes within the video using a Faster R-CNN module \cite{ren2016faster} that has been pre-trained \cite{anderson2018bottom} on the Visual Genome dataset \cite{krishna2017visual}. For multi-modality fusion, we set $C_1 = 512$ and $C_2 = 256$, resulting in a fused dimension of $C_3 = 768$, which supports varied feature dimensions as shown in Fig.~\ref{fig:feature_embedding}.

\subsection{Quantitative Comparison}
\subsubsection{Multi-person HOIs}
In the MPHOI-72 dataset, results in Table \ref{tab:MPHOI72} demonstrate $\modelshort$ not only surpasses the previous state-of-the-art models, ASSIGN \cite{morais2021learning} and 2G-GCN \cite{qiao2022geometric}, showcasing significant performance improvements, but also exhibits unparalleled stability. This is highlighted by $\modelshort$'s superior performance across all $\mathrm{F}{1}$ configurations coupled with substantially lower standard deviations. Specifically, in the $\mathrm{F}{1}@10$ score, $\modelshort$ achieves 71.3\%, which is approximately 3\% and 12\% higher than 2G-GCN and ASSIGN, respectively, marking a clear advancement in both predictive accuracy and consistency in the domain of human-object interaction recognition. These experimental outcomes further underscore the significance of geometric features in the multi-person Human-Object Interaction (MPHOI) domain. Models based solely on visual features, such as ASSIGN, are noticeably outperformed by those that incorporate both visual and geometric information. Although 2G-GCN integrates both visual and geometric features, its sub-optimal performance can be attributed to a lack of specificity in representing individual entities. Consequently, our model's superior performance and stability are not just a result of integrating multiple types of features but also our model's ability to specifically and effectively capture the nuanced dynamics of each entity involved in the interaction.

\begin{table}
\caption{Joined segmentation and label recognition on MPHOI-72.\label{tab:MPHOI72}}
\centering{}\resizebox{.53\textwidth}{!}{
\begin{tabular}{ccccc}
\toprule 
\multirow{2}{*}{Model} &  & \multicolumn{3}{c}{Sub-activity}\tabularnewline
\cmidrule{3-5}
 &  & $\mathrm{F}_{1}@10$ & $\mathrm{F}_{1}@25$ & $\mathrm{F}_{1}@50$\tabularnewline
\cmidrule{1-1} \cmidrule{3-5}
 ASSIGN \cite{morais2021learning} &  & 59.1 $\pm$ 12.1 & 51.0 $\pm$ 16.7 & 33.2 $\pm$ 14.0 \tabularnewline
2G-GCN \cite{qiao2022geometric} &  & 68.6 $\pm$ 10.4 & 60.8 $\pm$ 10.3 & 45.2 $\pm$ 6.5 \tabularnewline
\cmidrule{1-1} \cmidrule{3-5}
$\modelshort$ &  & \textbf{71.3} $\pm$ \textbf{5.0} & \textbf{65.8} $\pm$ \textbf{3.9} & \textbf{48.8} $\pm$ \textbf{5.3}\tabularnewline
\bottomrule
\end{tabular}}
\end{table}

\subsubsection{Single-person HOIs} 
In the CAD-120 dataset, as presented in Table \ref{tab:cad120}, $\modelshort$ demonstrates strong competitiveness in the single-person HOI scenarios. For both human sub-activity and object affordance labelling tasks, $\modelshort$ surpasses various prior methods, including those reliant on visual features like ATCRF\cite{koppula2016anticipating} and \cite{morais2021learning}, as well as the more sophisticated visual-geometric approach offered by 2G-GCN \cite{qiao2022geometric}. Notably, $\modelshort$ secures SOTA performance in both $\mathrm{F}{1}@10$ and $\mathrm{F}{1}@25$ metrics, registering improvements of 1.6\% and 0.1\% over ASSIGN and 2G-GCN, respectively. This achievement underscores $\modelshort$'s exceptional capability to accurately model and predict the dynamics of interactions, highlighting its adaptability and efficiency across different HOI challenges.

\begin{table}
\caption{Joined segmentation and label recognition on CAD-120.\label{tab:cad120}}
\centering{}\resizebox{.53\textwidth}{!}{
\begin{tabular}{ccccc}
\toprule 
\multirow{2}{*}{Model} &  & \multicolumn{3}{c}{Sub-activity}\tabularnewline
\cmidrule{3-5}
 &  & $\mathrm{F}_{1}@10$ & $\mathrm{F}_{1}@25$ & $\mathrm{F}_{1}@50$\tabularnewline
\cmidrule{1-1} \cmidrule{3-5}
rCRF \cite{sener2015rcrf} &  & 65.6 $\pm$ 3.2 & 61.5 $\pm$ 4.1 & 47.1 $\pm$ 4.3\tabularnewline
Independent BiRNN &  & 70.2 $\pm$ 5.5 & 64.1 $\pm$ 5.3 & 48.9 $\pm$ 6.8\tabularnewline
ATCRF \cite{koppula2016anticipating} &  & 72.0 $\pm$ 2.8 & 68.9 $\pm$ 3.6 & 53.5 $\pm$ 4.3\tabularnewline
Relational BiRNN &  & 79.2 $\pm$ 2.5 & 75.2 $\pm$ 3.5 & 62.5 $\pm$ 5.5\tabularnewline
ASSIGN \cite{morais2021learning} &  & 88.0 $\pm$ 1.8 & 84.8 $\pm$ 3.0 & 73.8 $\pm$ 5.8\tabularnewline
2G-GCN \cite{qiao2022geometric} &  & 89.5 $\pm$ 1.6  & 87.1 $\pm$ 1.8 & \textbf{76.2} $\pm$ 2.8\tabularnewline
\cmidrule{1-1} \cmidrule{3-5}
$\modelshort$ &  & \textbf{89.6} $\pm$ 2.1 & \textbf{87.3} $\pm$ \textbf{1.5} & 76.0 $\pm$ 3.5\tabularnewline
\bottomrule
\end{tabular}}
\end{table}

\vspace{-0.6cm}
\subsection{Qualitative Comparison}

In this section, we present a qualitative comparison of $\modelshort$ with the state-of-the-art method across the MPHOI-72 and CAD-120 datasets. 

\begin{figure}
\centering
\includegraphics[width=\linewidth]{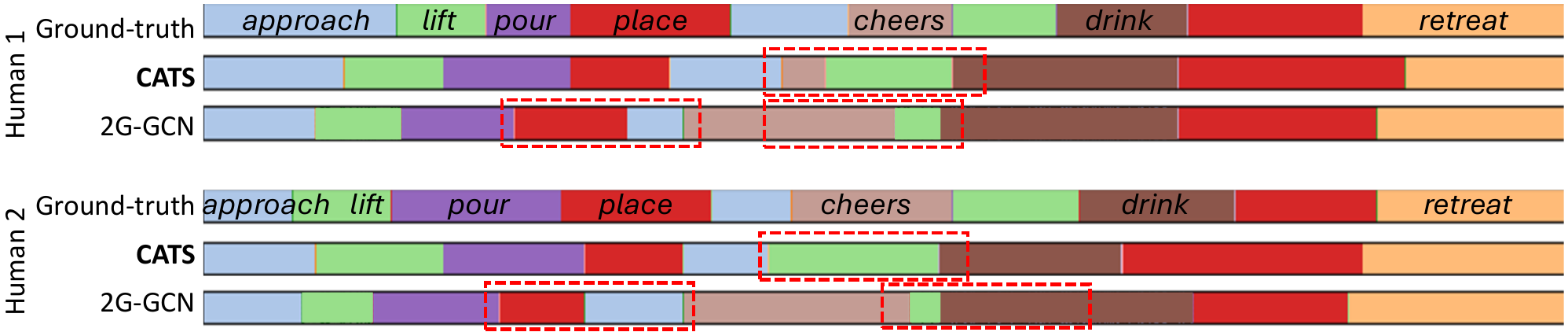}
\caption{Visualization of segmentation on MPHOI-72 for \textit{Cheering} activity. Red dashed boxes highlight major segmentation errors.}
\label{fig:vis_mp72_1}
\end{figure}

\begin{figure}
\centering
\includegraphics[width=\linewidth]{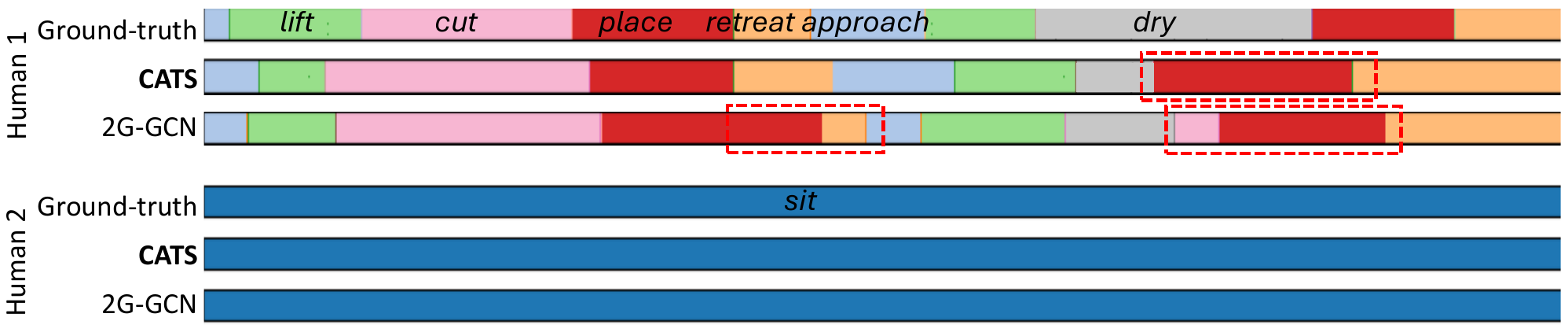}
\caption{Visualization of segmentation on MPHOI-72 for \textit{Hair cutting} activity. Red dashed boxes highlight major segmentation errors.}
\label{fig:vis_mp72_2}
\end{figure}

Fig.~\ref{fig:vis_mp72_1} and Fig.~\ref{fig:vis_mp72_2} illustrate \textit{Cheering} and \textit{Hair Cutting} activities within the MPHOI-72 dataset, comparing the segmentation and labeling tasks performed by $\modelshort$ and 2G-GCN \cite{qiao2022geometric} against the ground truth. Significant segmentation errors are marked with red dashed boxes. Although both methods exhibit some discrepancies in their predictions, $\modelshort$ more closely aligns with the ground truth, offering a more precise and stable visualization across a variety of actions. Conversely, 2G-GCN is prone to generating inappropriate sub-activities such as \textit{cheers} and \textit{lift} in the \textit{Cheering} activity. Moreover, in the \textit{Hair Cutting} activity, 2G-GCN inaccurately presents the \textit{cut} sub-activity into \textit{place} sub-activity, further deviating from the expected interaction dynamics. This comparison underscores the superior accuracy and reliability of $\modelshort$ in capturing and visualizing complex human-object interactions within diverse scenarios. 

\begin{figure}
\centering
\includegraphics[width=\linewidth]{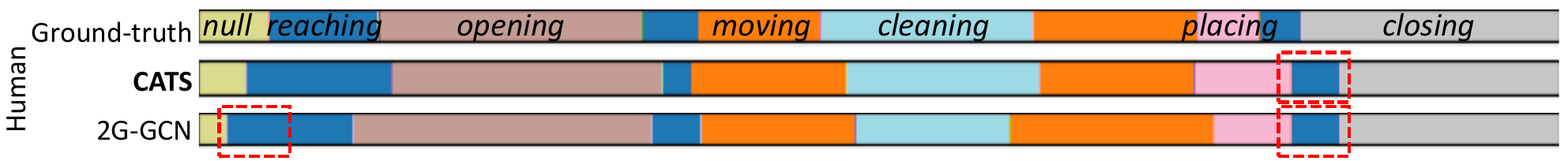}
\caption{Visualization of segmentation on CAD-120 for \textit{Cleaning objects} activity. Red dashed boxes highlight major segmentation errors.}
\label{fig:cad_vis1}
\end{figure}

\begin{figure}
\centering
\includegraphics[width=\linewidth]{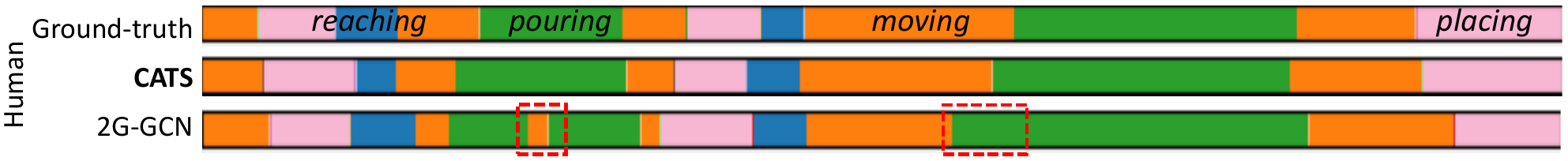}
\caption{Visualization of segmentation on CAD-120 for \textit{Making Cereal} activity. Red dashed boxes highlight major segmentation errors.}
\label{fig:cad_vis2}
\end{figure}

Fig.~\ref{fig:cad_vis1} and Fig.~\ref{fig:cad_vis2} illustrate the \textit{Cleaning Objects} and \textit{Making Cereal} activities from the single-person CAD-120 dataset, with abnormal segmentation instances accentuated by red dashed boxes. For the \textit{Cleaning Objects} activity, both methods effectively match the overall ground truth. However, $\modelshort$ provides a visualization that more closely approximates the ground truth. In the \textit{Making Cereal} activity, $\modelshort$ significantly outperforms 2G-GCN, particularly in sub-activities such as \textit{pouring}, \textit{moving}, and \textit{reaching}, while 2G-GCN yields some inaccurate segmentations. The enhanced precision of $\modelshort$ in capturing the intricacies of each activity highlights its superior performance, excelling in the identification and precise representation of detailed actions and interactions within the scenes, thus delivering a more accurate and reliable analysis of the activities performed.

\vspace{-0.2cm}
\subsection{Alternative Architectures and Ablation Studies}

\subsubsection{Architecture Alternatives Comparison}

We evaluate the HOI recognition performance on the MPHOI-72 and CAD-120 datasets by conducting tests on various alternative model structures. The experimental outcomes, as detailed in Tables \ref{tab:alter_MPHOI72} and \ref{tab:alter_CAD120}, reveal that our model consistently delivers superior results compared to these alternatives. This superior performance is likely attributable to the unique consideration our model gives to category-level interactions, specifically the distinct analysis of human-human and object-object interactions. Unlike other approaches that might treat interactions generically or overlook the nuanced distinctions between different types of interactions, our model maintains a comprehensive view. 

\begin{table}[htbp]
    \centering
    \begin{minipage}[t]{0.49\linewidth}
        \centering
        \renewcommand{\arraystretch}{1.05} 
        \caption{\textls[-1]{Comparison between architecture alternatives and $\modelshort$ on MPHOI-72.}}
        \label{tab:alter_MPHOI72}
        \begin{adjustbox}{width=\linewidth}
        \begin{tabular}{ccccc}
        \toprule 
        \multirow{2}{*}{Model} &  & \multicolumn{3}{c}{Sub-activity}\tabularnewline
        \cmidrule{3-5}
         &  & $\mathrm{F}_{1}@10$ & $\mathrm{F}_{1}@25$ & $\mathrm{F}_{1}@50$\tabularnewline
        \cmidrule{1-1} \cmidrule{3-5}
        Independent-entity architecture &  & 65.1 $\pm$ 3.3 & 58.7 $\pm$ 1.7 & 40.4 $\pm$ 3.9 \tabularnewline
        2G-GCN \cite{qiao2022geometric} &  & 68.6 $\pm$ 10.4 & 60.8 $\pm$ 10.3 & 45.2 $\pm$ 6.5 \tabularnewline
        \cmidrule{1-1} \cmidrule{3-5}
        $\modelshort$ &  & 71.3 $\pm$ 5.0 & 65.8 $\pm$ 3.9 & 48.8 $\pm$ 5.3\tabularnewline
        \bottomrule
        \end{tabular}
        \end{adjustbox}
    \end{minipage}\hfill
    \begin{minipage}[t]{0.49\linewidth}
        \centering
        \renewcommand{\arraystretch}{1} 
        \caption{Comparison between architecture alternatives and $\modelshort$ on CAD-120.}
        \label{tab:alter_CAD120}
        \begin{adjustbox}{width=\linewidth}
        \begin{tabular}{ccccc}
        \toprule 
        \multirow{2}{*}{Model} &  & \multicolumn{3}{c}{Sub-activity}\tabularnewline
        \cmidrule{3-5}
         &  & $\mathrm{F}_{1}@10$ & $\mathrm{F}_{1}@25$ & $\mathrm{F}_{1}@50$\tabularnewline
        \cmidrule{1-1} \cmidrule{3-5}
        Independent-entity architecture &  & 85.9 $\pm$ 4.0 & 84.1 $\pm$ 4.9 & 72.8 $\pm$ 5.2 \tabularnewline
        2G-GCN \cite{qiao2022geometric} &  & 89.5 $\pm$ 1.6  & 87.1 $\pm$ 1.8 & 76.2 $\pm$ 2.8\tabularnewline
        \cmidrule{1-1} \cmidrule{3-5}
        $\modelshort$ &  & 89.6 $\pm$ 2.1 & 87.3 $\pm$ 1.5 & 76.0 $\pm$ 3.5\tabularnewline
        \bottomrule
        \end{tabular}
        \end{adjustbox}
    \end{minipage}
\end{table}

\vspace{-0.5cm}
\begin{table}
\caption{Results of different GCN layers in multi-category multi-modality fusion on MPHOI-72.\label{tab:nlayer_gcn}}
\centering{}\resizebox{.6\textwidth}{!}{
\begin{tabular}{ccccc}
\toprule 
\multirow{2}{*}{Model} &  & \multicolumn{3}{c}{Sub-activity}\tabularnewline
\cmidrule{3-5}
 &  & $\mathrm{F}_{1}@10$ & $\mathrm{F}_{1}@25$ & $\mathrm{F}_{1}@50$\tabularnewline
\cmidrule{1-1} \cmidrule{3-5}
1-layer GCN &  & 70.4 $\pm$ 1.7 & 62.0 $\pm$ 2.5 & 43.9 $\pm$ 3.8 \tabularnewline
2-layer GCN &  & 68.8 $\pm$ 4.3 & 62.1 $\pm$ 4.3 & 44.0 $\pm$ 3.3 \tabularnewline
3-layer GCN &  & 67.4 $\pm$ 4.2 & 63.3 $\pm$ 3.4 & 44.2 $\pm$ 1.3 \tabularnewline
5-layer GCN &  & 70.4 $\pm$ 5.7 & 60.0 $\pm$ 2.3 & 43.7 $\pm$ 2.2 \tabularnewline
\cmidrule{1-1} \cmidrule{3-5}
4-layer GCN (Ours) &  & \textbf{71.3} $\pm$ \textbf{5.0} & \textbf{65.8} $\pm$ \textbf{3.9} & \textbf{48.8} $\pm$ \textbf{5.3}\tabularnewline
\bottomrule
\end{tabular}}
\end{table}

\vspace{-0.5cm}
\subsubsection{GCN Layers for Geometric Feature Learning} In this section, we conduct ablation studies to elucidate the impact of the depth of GCN layers on the geometric learning of human joints and object keypoints within our network, results are shown in Table \ref{tab:nlayer_gcn}. To assess the influence of GCN layer depth on model performance, we explore configurations with 1, 2, 3, 4, and 5 GCN layers. Through this comparative analysis, we aim to identify the most effective layer depth that balances computational efficiency with the nuanced understanding of spatial relationships essential for interpreting complex interactions between humans and objects. The results indicate that a configuration of 4-layer GCN offers the optimal balance, providing the best performance in terms of both accuracy and computational efficiency. This depth allows for sufficient complexity to understand and model the geometric relationships critical for accurate interaction recognition, without incurring the diminishing returns or increased computational demand associated with additional layers.

\section{Conclusion}
In conclusion, we propose $\modelshort$, an advanced end-to-end framework that enhances video-based HOI recognition through sophisticated integration of category and scenery level analyses. It first fuses multi-modal features of different categories, and then construct a scenery interactive graph to learn the relationships between these categories. $\modelshort$ demonstrates superior performance on key benchmarks such as MPHOI-72 and CAD-120 datasets, showcasing the effectiveness of multi-person and single-person HOI recognition.

\section*{Acknowledgement}
This research is supported in part by the EPSRC NortHFutures project (ref: EP/X031012/1).

%
%
\bibliographystyle{splncs04}
\bibliography{abbr,main}

\end{document}